%% file: ASE2018.tex
\documentclass[sigconf, screen]{acmart}
\usepackage{amsmath}
\usepackage{algorithm}
\usepackage{algpseudocode}
\usepackage{tikz} 
\usetikzlibrary{shapes,arrows}
\usepackage{pgfplots}
\pgfplotsset{compat=1.14}
\usepackage{float}
\usepackage{booktabs}
\usepackage{multirow}
\usepackage{graphicx}
\usepackage{makecell}
\usepackage{tcolorbox,xspace,amssymb,amsmath,amsthm}
\usepackage{balance}

\usepackage{fp}
\usepackage{color}
\usepackage{colortbl}

\definecolor{LightGreen}{rgb}{0.7, 1.0, 0.7}
\definecolor{LightRed}{rgb}{1.0, 0.7, 0.7}
\definecolor{LightBlue}{rgb}{0.7, 0.7, 1.0}
\definecolor{TableHeader}{rgb}{0.9, 0.9, 0.9}

\newcommand{\FT}{\textsc{Aequitas}\xspace}

\setcopyright{acmcopyright}
\acmPrice{15.00}
\acmDOI{10.1145/3238147.3238165}
\acmYear{2018}
\copyrightyear{2018}
\acmISBN{978-1-4503-5937-5/18/09}
\acmConference[ASE '18]{Proceedings of the 2018 33rd ACM/IEEE International Conference on Automated Software Engineering}{September 3--7, 2018}{Montpellier, France}
\acmBooktitle{Proceedings of the 2018 33rd ACM/IEEE International Conference on Automated Software Engineering (ASE '18), September 3--7, 2018, Montpellier, France}

\begin{document}



\title{Automated Directed Fairness Testing}


\author{Sakshi Udeshi}
\affiliation{%
  \institution{Singapore Univ. of Tech. and Design}
  \country{Singapore}
}
\email{sakshi_udeshi@mymail.sutd.edu.sg}

\author{Pryanshu Arora}
\affiliation{%
  \institution{BITS Pilani}
  \country{India}
}
\email{pryanshu23@gmail.com}

\author{Sudipta Chattopadhyay}
\affiliation{%
  \institution{Singapore Univ. of Tech. and Design}
  \country{Singapore}
}
\email{sudipta_chattopadhyay@sutd.edu.sg}

\input{abstract}

\maketitle

\input{introduction}

\input{overview}

\input{approach}

\input{results}

\input{relatedWork}

\input{threatsToValidity}

\input{conclusion}

\section*{Acknowledgment}
The authors would like to thank Chundong Wang and the anonymous reviewers 
for their insightful comments. The first author is supported by the President's 
Graduate Fellowship funded by the Ministry of Education, Singapore.

\newpage

\balance
\bibliographystyle{plainurl}
\bibliography{ASE2018}

\end{document}

%% file: abstract.tex
\begin{abstract}

Fairness is a critical trait in decision making. As machine-learning models 
are increasingly being used in sensitive application domains (e.g. education 
and employment) for decision making, it is crucial that the decisions computed 
by such models are free of unintended bias. But how can we automatically 
validate the fairness of arbitrary machine-learning models? 
For a given machine-learning model and a set of sensitive input parameters, 
our \FT approach automatically discovers discriminatory inputs that highlight 
fairness violation. At the core of \FT are three novel strategies to employ 
probabilistic search over the input space with the objective of uncovering 
fairness violation. Our \FT approach leverages inherent robustness property 
in common machine-learning models to design and implement scalable test 
generation methodologies. An appealing feature of our generated test inputs 
is that they can be systematically added to the training set of the underlying 
model and improve its fairness. To this end, we design a fully automated module 
that guarantees to improve the fairness of the model. 

We implemented \FT and we have evaluated it on six state-of-the-art classifiers. 
Our subjects also include a classifier that was designed with fairness in mind. 
We show that \FT effectively generates inputs to uncover fairness violation in all 
the subject classifiers and systematically improves the fairness of respective 
models using 
the generated test inputs. In our evaluation, \FT generates up to 70\% 
discriminatory inputs (w.r.t. the total number of inputs generated) and leverages 
these inputs to improve the fairness up to 94\%.    

\end{abstract}

%% file: introduction.tex
\section{Introduction}
\label{sec:introduction}

Nondiscrimination is one of the most critical factors for social protection 
and equal human rights. The basic idea behind non-discrimination is to 
eliminate any societal bias based on sensitive attributes, such as race, 
gender or religion. For example, it is not uncommon to discover the 
declaration of following non-discrimination policy in 
universities~\cite{umich-site}:
\begin{verse}
{\em``The University is committed to a policy of equal opportunity for all 
persons and does not discriminate on the basis of race, color, national 
origin, age, marital status, sex, sexual orientation, gender identity, gender 
expression, disability, religion, height, weight, or veteran status in 
employment, educational programs and activities, and admissions"}
\end{verse}
Due to the massive progress in machine learning in the last few decades, its 
application has now escalated over a variety of sensitive domains, including 
education and employment. The key insight is to primarily automate decision 
making via machine-learning models. On the flip side, such models may 
introduce unintended societal bias due to the presence of bias in their training 
dataset. 
This, in turn, violates the non-discrimination policy that the respective 
organization or the nation is intended to fight for. The validation of 
machine-learning models, to check for possible discrimination, is therefore 
critically important. 

\begin{figure}[t]
\begin{center}
\begin{tabular}{c}
\rotatebox{0}{
\includegraphics[scale = 0.7]{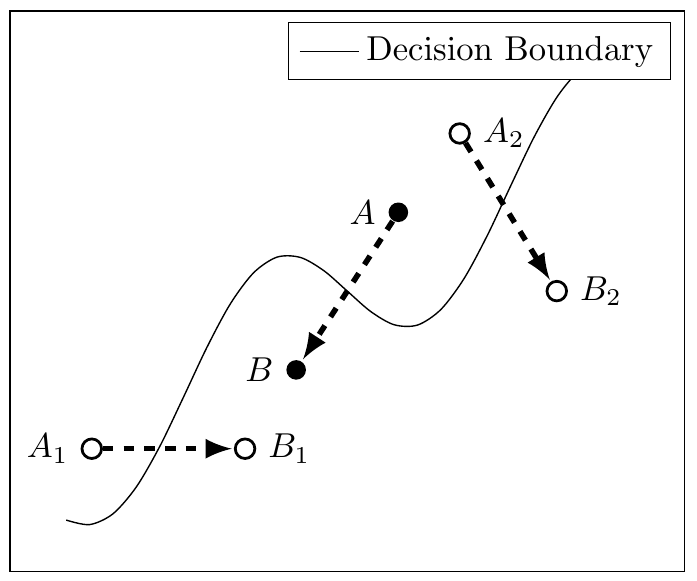}}
\end{tabular}
\end{center}
\vspace*{-0.1in}
\caption{Classifier fairness}
\label{fig:fairness-basic}
\vspace*{-0.2in}
\end{figure}

In this paper, we are concerned about the case that any two 
individuals who are similar with respect to a job at hand should also 
be treated in a similar fashion during decision making. Thus, we 
focus towards {\em individual fairness}, as it is critical for 
eliminating societal bias and aim to check for discrimination that 
might violate {\em individual fairness}~\cite{fairness_awareness_dwork}. 
The precise nature of such discrimination depends on the machine-learning 
model and its input features. Consequently, given a machine-learning model 
and the input features of the model, it is possible to systematically 
explore the input space and discover inputs that induce discrimination. 
We call such inputs {\em discriminatory inputs}. The primary objective 
of this paper is to design scalable techniques that facilitate rapid 
discovery of discriminatory inputs. In particular, given a machine-learning 
model and a set of discriminatory input features (e.g. race, religion, etc.), 
our \FT approach {\em automatically discovers inputs to clearly highlight 
the discriminatory nature of the model under test}.

As an example, consider the decision boundary of a classifier shown in 
\autoref{fig:fairness-basic}. Assume the two points $\mathbf{A}$ and 
$\mathbf{B}$ that differ only in being {\em $\mathit{Gender}_A$} or 
{\em $\mathit{Gender}_B$}.
Despite being vastly similar, except in the gender aspect, the model classifies 
the points $\mathbf{A}$ and $\mathbf{B}$ differently. If we consider that 
such a classifier is used to predict the level of salary, then it certainly 
introduces unintended societal bias based on gender. 
Such unfair social biases not only affect the decisions of today but also 
might amplify it for future generations.
The reason behind the discrimination (i.e. unfairness), as shown between 
points $\mathbf{A}$ and $\mathbf{B}$, can be due to outdated training data 
that unintentionally introduces bias in certain attributes of the classifier 
model, e.g., gender in \autoref{fig:fairness-basic}. Using our \FT approach, 
we automatically discover the existence of inputs similar to $\mathbf{A}$ 
and $\mathbf{B}$ with high probabilities. These inputs, then, are used to 
systematically retrain the model and reduce its unfairness.


The reason \FT works is due to its directed strategy for test generation. 
In particular, \FT exploits the inherent robustness
property of common machine learning models for systematically directing test 
generation. As a result of this robustness property, the models 
should exhibit low variation in 
their output(s) with small perturbations in their input(s). 
For example, consider the points $\mathbf{A}_1$ and $\mathbf{A}_2$ which are 
in the neighbourhood of the point $\mathbf{A}$. Since the point $\mathbf{A}$ 
exhibits discriminatory nature, it is likely that both points 
$\mathbf{A}_1$ and $\mathbf{A}_2$ will be discriminatory, as reflected 
via the presence of points $\mathbf{B}_1$ and $\mathbf{B}_2$, respectively. 
In our \FT approach, we first randomly sample the input space to discover 
the presence of discriminatory inputs (e.g. point $\mathbf{A}$ in 
\autoref{fig:fairness-basic}). Then, we search the neighbourhood of these 
inputs, as discovered during the random sampling, to find the presence of 
more inputs (e.g. points $\mathbf{A}_1$ and $\mathbf{A}_2$ in 
\autoref{fig:fairness-basic}) of the same nature.

An appealing feature of \FT is that it leverages the generated test inputs 
and systematically retrains the machine-learning model under test to reduce 
its unfairness. The retraining module is completely automatic and it  
therefore acts as a significant aid to the software engineers to improve 
the (individual) fairness of machine-learning models. 
The directed test generation and automated retraining set \FT apart from 
the state-of-the-art in fairness testing~\cite{fairness_fse_2017}. 
While existing work~\cite{fairness_fse_2017} also considers test 
generation, such tests were generated randomly. If the discriminatory 
inputs are located only in specialized locations of the input space, then 
random test generators are unlikely to be effective in finding individuals 
discriminated by the corresponding model. To this end, \FT empirically 
validates that a directed test generation, to uncover the discriminatory 
input regions, is indeed more desirable than random test generation. 
Moreover, \FT provides statistical evidence that if it fails to discover 
any discriminatory input, then the machine-learning model under test is 
fair with high probability. 

The remainder of the paper is organized as follows. After providing an 
overview of \FT (\autoref{sec:overview}), we make the following contributions: 

\vspace{-0.25pc}
\begin{enumerate}

\item We present \FT, a novel approach to systematically generate discriminatory 
test inputs and uncover the fairness violation in machine-learning models. To 
this end, we propose three different strategies with varying levels of 
complexity (\autoref{sec:approach}).

\item We present a fully automated technique to leverage the generated 
discriminatory inputs and systematically retrain the machine-learning models to 
improve its fairness (\autoref{sec:approach}). 

\item We provide an implementation of \FT based on python. Our implementation and 
all experimental data are publicly available (\autoref{sec:results}). 

\item We evaluate our \FT approach with six state-of-the-art classifiers including 
a classifier that was designed with fairness in mind. Our evaluation reveals that 
\FT is effective in generating discriminatory inputs and improving the fairness of 
the classifiers under test. In particular, \FT generated up to 70\% discriminatory 
inputs (w.r.t. the total number of inputs generated) and improved the fairness up 
to 94\% (\autoref{sec:results}).

\end{enumerate}

After discussing the related work (\autoref{sec:relatedWork}), we outline different 
threats to validity (\autoref{sec:threats}) before conclusion and consequences 
(\autoref{sec:conclusion}).

%% file: overview.tex
\section{Background}
\label{sec:overview}
In this section, we will discuss the critical importance of fairness testing and outline 
the key insight behind our approach. 


\smallskip\noindent
\textbf{Importance of fairness}
The usage of machine learning is increasingly being observed in areas that are under the 
purview of anti-discrimination laws. In particular, application domains such as law 
enforcement, credit, education and employment can all benefit from machine learning. 
Hence, it is crucial that decisions influenced by any machine-learning model are free of 
any unnecessary bias.

As an example, consider a machine-learning model that predicts the income levels of a 
person. It is possible that such a model was trained on a dataset, which, in turn was 
unfairly biased to a certain gender or a certain race. As a result, for all equivalent 
characteristics, barring the gender or race, the credit worthiness of a person will be 
predicted differently by this model. If financial institutions used such a 
model to determine the credit worthiness of an individual, then individuals might be 
disqualified only on the basis of their gender or race. Such a discrimination is 
certainly undesirable, as it reinforces and amplifies the unfair biases that we, as a 
society are continuously fighting against. 


\smallskip\noindent
\textbf{Fairness in \FT}
\FT aims to discover the violation of {\em individual fairness}~\cite{fairness_awareness_dwork} 
in machine-learning models. This means, \FT aims to find instances of pair of inputs 
$I$ and $I'$ that are classified differently despite being vastly similar. 
The similarity between inputs $I$ and $I'$ is based on a set of potentially 
discriminatory input parameters (see Definition~\ref{def:discriminatory}). 
Detecting the violation of individual fairness is challenging. This is because  
inputs that are prone to the violation of individual fairness might be located 
only in specific regions of the input space of a model. Consequently, specialized 
and directed techniques are required to rapidly locate these input regions. 
This is the primary motivation behind the development of \FT. For the rest of 
the paper, we will simply use the term fairness (instead of individual fairness) 
in the light of our \FT approach (see Definition~\ref{def:discriminatory}).


\smallskip\noindent
\textbf{Towards fair machine-learning models}
A naive approach to design {\em fair} machine-learning models is to ignore certain 
sensitive attributes such as race, color, religion, gender, disability, or family 
status. It is natural to assume that if such attributes are held back from decision 
making, then the respective model will not discriminate. Unfortunately, such an approach 
of accomplishing fairness through {\em blindness} fails. This is because of the presence 
of redundant encoding in the training dataset~\cite{pedreshi2008discrimination}. 
Due to the redundant encoding, it is frequently possible to predict 
the unknown (sensitive) attributes from other seemingly innocuous features. 
For example, consider certain ethnic groups in a city that are geographically bound 
to certain areas. In such cases, even if a machine-learning model in a financial 
institute does not use ethnicity as a parameter to decide credit worthiness, it is 
possible to guess 
ethnicity from geographic locations, which indeed might be a 
parameter for the model. Therefore, it is critical to systematically test a 
machine-learning model to validate its fairness property.


\smallskip\noindent
\textbf{Why fairness testing is different}
In contrast to classic software testing, testing machine-learning models face additional 
challenges. Typically, these models are deployed in contexts where the formal specification 
of the software functionality is difficult to develop. In fact, such models are 
designed to learn from existing data because of the challenges in creating a mathematical 
definition of the desired software properties. 
Moreover, an erroneous software behaviour can be rectified by retraining the machine-learning 
models. However, for classic software, a software bug is typically fixed via modifying 
the responsible code.
%

\begin{figure*}[t]
\begin{center}
\begin{tabular}{cc}
\rotatebox{0}{
\includegraphics[scale = 0.7]{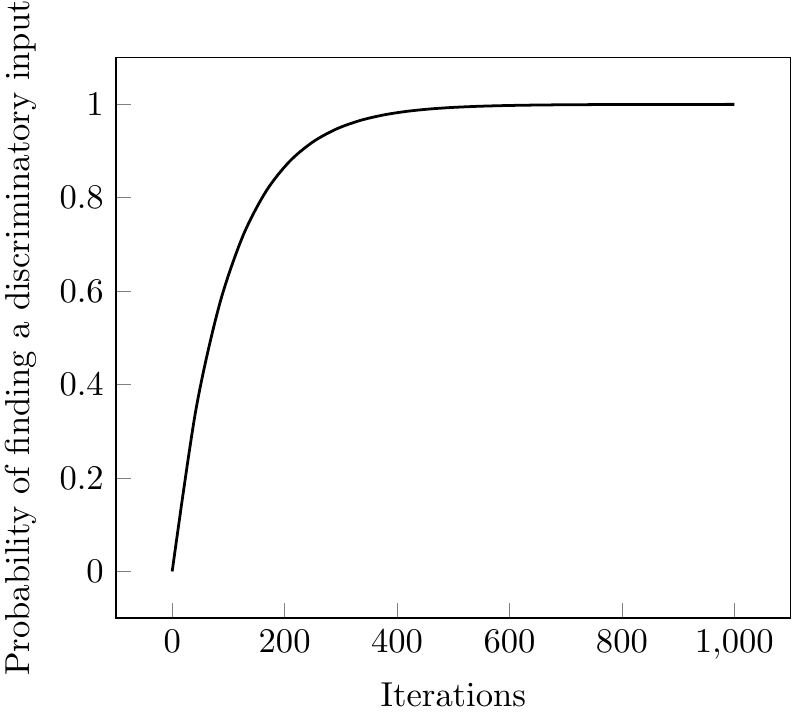}} & 
\rotatebox{0}{
\includegraphics[scale = 0.44]{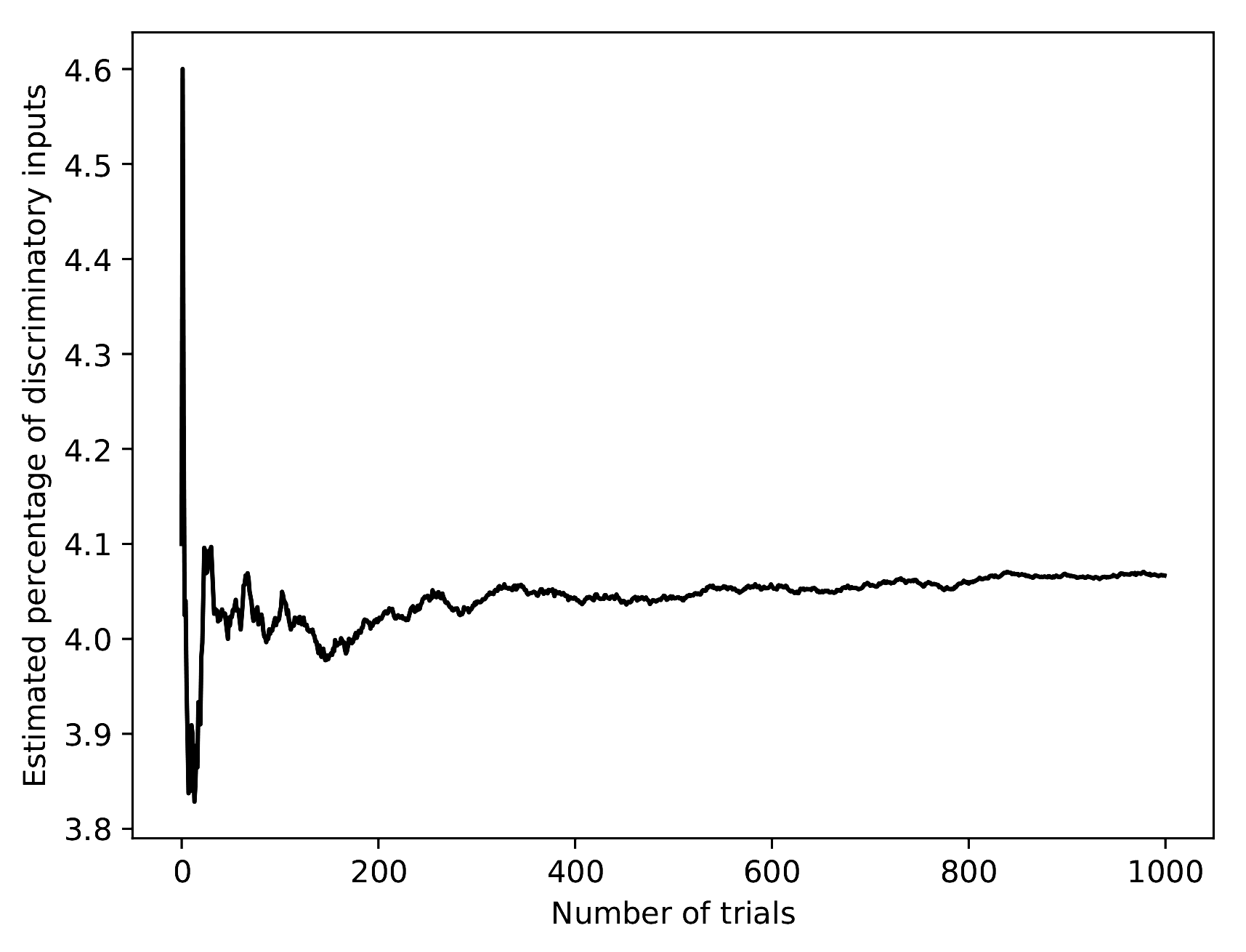}} \\
\textbf{(a)} & \textbf{(b)}\\
\end{tabular}
\end{center}
\vspace*{-0.2in}
\caption{(a) Probability of finding discriminatory inputs, (b) Estimation of the percentage of discriminatory inputs}
\label{fig:example}
\end{figure*}


\smallskip\noindent
\textbf{State-of-the-art in fairness testing}
The state-of-the-art in systematic testing of software fairness 
is still at its infancy. In contrast to existing work~\cite{fairness_fse_2017}, 
\FT focuses on directed test generation strategy. As evidenced by 
our evaluation, this is crucial to locate specific input regions 
that violate individual fairness.  
%
%
To illustrate our objective, consider a machine-learning model $f$ and its inputs $I$ and 
$I'$. $I$ differs from $I'$ only in being assigned a different value in a potentially discriminatory 
input parameter. For example, if {\em gender} is the potentially discriminatory input parameter, 
then $I$ will be different from $I'$ only in being $\mathit{Gender}_A$ or $\mathit{Gender}_B$.
We are interested to discover inputs $I$ or $I'$, where the difference in outputs of the model, 
captured via $\left | f(I) - f(I') \right |$, is beyond a pre-determined 
threshold. We call such inputs $I$ or $I'$ to be {\em discriminatory inputs} for 
the model $f$. It is important to note that the discrimination threshold and the potentially 
discriminatory input parameters are supplied by the users of our tool. In the preceding 
example, the potentially discriminatory input parameter, i.e., {\em gender} can be 
specified by the user. Similarly, users can also fine tune the value at which 
$\left | f(I) - f(I') \right |$ is considered to be discriminatory.



\smallskip\noindent
\textbf {Robustness in machine learning}
Robustness is a notion that says that the output of a machine-learning 
model is not dramatically affected by small changes to its input~\cite{robustness_classifier}. 
Assume a model $f$, let $i$ be the input to $f$ and $\delta$ be a small value. 
If $f$ is robust, then $f(i) \approx f(i + \delta)$. 
Nevertheless, existing techniques provide evidence to find inputs that violate this 
robustness property. Such inputs are called adversarial 
inputs~\cite{adversarial_examples}~\cite{adversarial_perturbations}~\cite{attack_machine_learning}. 
However, adversarial inputs generally cover only a small fraction of the entire input space. 
This is evident by the fact that adversarial inputs need to be crafted using very specialized 
techniques. 
Additionally, \FT is designed to avoid these adversarial input regions by systematically 
directing the test generators. Intuitively, \FT achieves this by reducing the 
probability to explore an input region when tested inputs from the region did not exhibit 
discriminatory nature (see Algorithm \autoref{algo:local} for details). Consequently, 
if adversarial or non-robust input regions do not exhibit discriminatory nature, such 
regions will eventually be explored only with very low probability.



\begin{figure}[t]
\begin{center}
\begin{tabular}{c}
\rotatebox{0}{
\includegraphics[scale = 0.41]{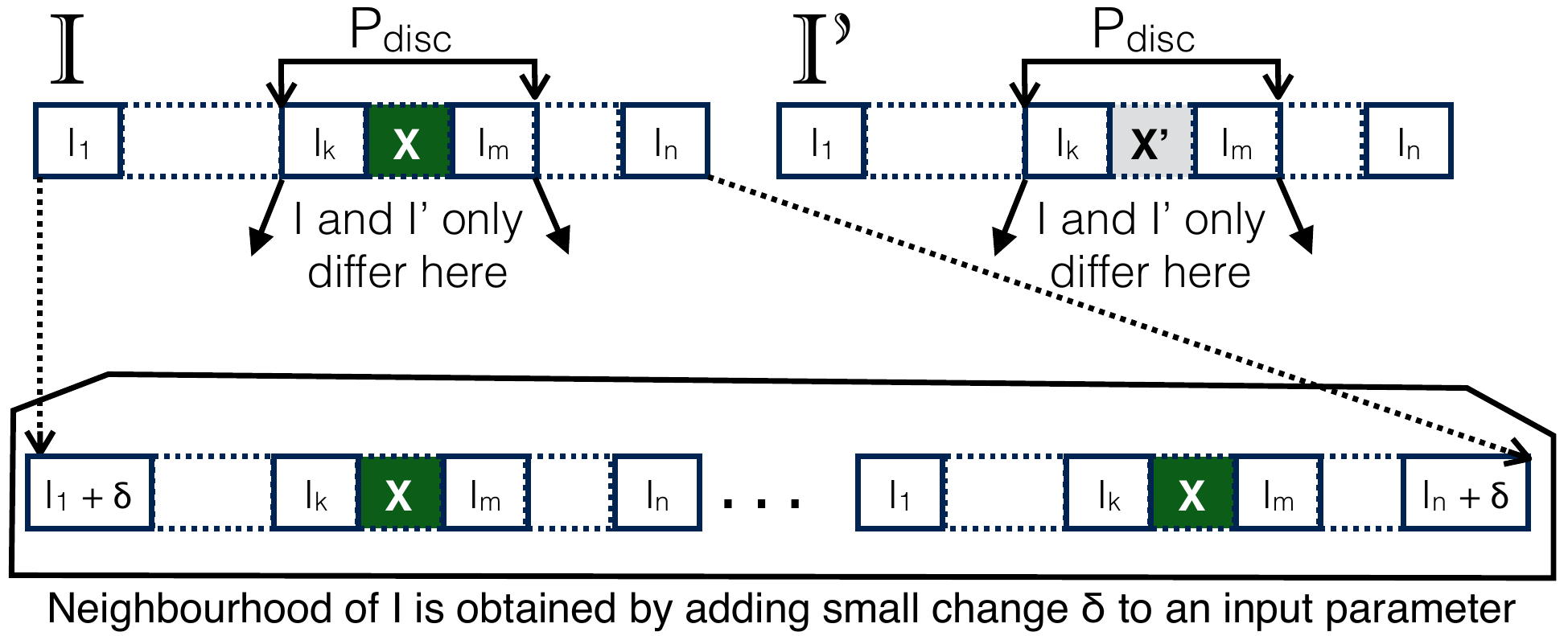}}
\end{tabular}
\end{center}
\vspace*{-0.1in}
\caption{Our \FT approach at a glance}
\label{fig:aq-overview}
\vspace*{-0.2in}
\end{figure}

\section{Approach at a glance}

We propose, design and evaluate three schemes, with varying levels of complexities, 
to systematically uncover software fairness problems. The crucial components of 
our approach are outlined below. 



\smallskip\noindent
\textbf{Global search}
In the first step of all our proposed 
schemes, we uniformly sample the inputs and record the discriminatory inputs that 
we find. In the light of uniformly sampling the input space, we can guarantee, with very high 
probability, to discover a discriminatory input, if such an input exists. 
For instance, \autoref{fig:example}(a) highlights the probability of finding a discriminatory 
input in an input space with only 1\% discriminatory inputs. Therefore, if discriminatory 
inputs exist, the first step of our proposed schemes guarantee to find at least one such 
input with high probabilities. 


\smallskip\noindent
\textbf{Local search}
The second step of our proposed schemes share the following hypothesis: {\em If there exists 
a discriminatory input $I \in \mathbb{I}$, where $\mathbb{I}$ captures the input domain, then 
there exist more discriminatory inputs in the input space closer to $I$}. 
The input domain $\mathbb{I}$ can be considered as the cartesian product of the domain 
of $n$ input parameters, say $P_1,P_2, \ldots, P_n$. We assume $\mathbb{I}_k$ captures 
the domain of input parameter $P_k$.
Therefore, $\mathbb{I} = \mathbb{I}_1 \times \mathbb{I}_2 \times \ldots \times \mathbb{I}_n$. 
An input parameter $p \in \bigcup_{i=1}^{n} P_i$ can be potentially discriminatory if the 
output of the machine-learning model should not be biased towards specific values in 
$\mathbb{I}_p$. Without loss of generality, we assume a subset of parameters 
$P_{disc} \subseteq \bigcup_{i=1}^{n} P_i$ to be potentially discriminatory. For an input 
$I \in \mathbb{I}$, we use $I_k$ to capture the value of parameter $P_k$ within 
input $I$. Based on this notion, we explore the following methods to realize our 
hypothesis. Our methods differ on how we systematically explore the neighbourhood 
of a discriminatory input $I^{(d)}$. $I^{(d)}$, in turn, was discovered in the 
first step of \FT. 

\begin{enumerate}
\item 
First a parameter $p \in  \bigcup_{i=1}^{n} P_i \setminus P_{disc}$ 
is randomly chosen. Then a small perturbation (i.e. change) $\delta$ is added 
to $I^{(d)}_p$. Typically $\delta \in \{-1,+1\}$ as we consider integer and 
real-valued input parameters in our evaluation.   


\item In the second method, we assign probabilities on how to perturb a chosen 
parameter. A specific parameter $p \in \bigcup_{i=1}^{n} P_i \setminus P_{disc}$ 
is still chosen uniformly at random. However, if a given perturbation $\delta$ of $I^{(d)}_p$ 
consistently yields discriminatory inputs, then the perturbation $\delta$ is 
employed with higher probability. Since $\delta$ typically belongs to a small 
set of values, such a strategy works efficiently in practice.  

\item The third method augments the second method by refining probabilities to perturb 
an input parameter. Concretely, if perturbing the value of parameter 
$p \in \bigcup_{i=1}^{n} P_i \setminus P_{disc}$ consistently yields discriminatory inputs, 
then the parameter $p$ will be significantly more likely to be chosen for perturbation.
\end{enumerate}
Our proposed methodologies are fully automated, they do not require the source code 
of the models and work efficiently in practice for state-of-the-art classifiers. 
%

\autoref{fig:aq-overview} illustrates \FT approach when $I$ and $I'$ 
were discovered in the first step. Then, the second step explored the 
neighbourhood of $I$ by adding small changes $\delta$ to an input parameter. 


\smallskip\noindent
\textbf{Estimation of discriminatory inputs}
An appealing feature of \FT is that we can estimate the percentage 
of discriminatory inputs in $\mathbb{I}$. To this end, we leverage the law of 
large numbers (LLN) in probability theory. In particular, we generate $K$ inputs 
uniformly at random and check whether they can lead to discriminatory inputs. 
Assume that $K' \leq K$ inputs turn out to be discriminatory. We compute the 
ratio $\frac{K'}{K}$ over a large number of trials. According to LLN, the 
average of these ratios closely approximates the actual percentage of discriminatory 
inputs in $\mathbb{I}$. \autoref{fig:example}(b) highlights such convergence after 
only 400 trials when $K$ was chosen to be 1000.





\smallskip\noindent
\textbf{Why \FT works?}
The reason \FT works is because of the robustness property of common 
machine-learning models. 
In particular, if we perturb the input to a model by some small $\delta$, 
then the output is not expected to change dramatically. As we expect the 
machine-learning models under test to be relatively robust, we can leverage 
their inherent robustness property to systematically generate test inputs that 
exhibit similar characteristics. In our \FT approach, we focus on the discriminatory 
nature of a given input. We aim to discover more discriminatory inputs in the 
proximity of an already discovered discriminatory input leveraging the 
robustness property.



\smallskip\noindent
\textbf{How \FT can be used to improve software fairness?}
We have designed a fully automated module that leverages on the discriminatory inputs 
generated by \FT and retrains the machine-learning model under test. We empirically 
show that such a strategy provides useful capabilities to a developer. Specifically, 
our \FT approach automatically improves the fairness of machine-learning models 
via retraining. For instance, in certain decision tree classifiers, our \FT approach 
reduced the fraction of discriminatory inputs up to 94\%.






%% file: approach.tex
%

\section{Detailed approach}
\label{sec:approach}

In this section, we discuss our \FT approach in detail. To this end, we will use the 
notations captured in \autoref{tab:notation}. 
\begin{table}[t]
	\centering
	\caption{Notations used in \FT approach}
	\vspace*{-0.1in}
	\label{tab:notation}
	\begin{tabular}{| c | p{7cm} | }
	\hline
	$n$      & The number of input parameters to the machine-learning model under test   \\
	\hline
	$\mathbb{I}$ & The input domain of the model\\
	\hline
	$P_i$ & The $i$-th input parameter of the model\\
	\hline
	$P$ & Set of all input parameters, i.e., $P = \bigcup_{i=1}^{n} P_i$\\
	\hline
	$P_{disc}$ & Set of sensitive or potentially discriminatory input parameters (e.g. gender). 
Clearly, $P_{disc} \subseteq \bigcup_{i=1}^{n} P_i$\\
	\hline
	$I_{p}$ & The value of input parameter $p$ in input $I \in \mathbb{I}$\\
	\hline
	$\gamma$ & A pre-determined discrimination threshold\\
	\hline
\end{tabular}
	\vspace*{-0.15in}
\end{table}

Our approach revolves around discovering {\em discriminatory inputs} via systematic 
{\em perturbation}. We introduce the notion of discriminatory inputs and perturbation 
formally before delving into the algorithmic details of our approach.

\theoremstyle{definition}
\begin{definition}{\textbf{(Discriminatory Input and fairness)}}
\label{def:discriminatory}
{
Let $f$ be a classifier under test, $\gamma$ be the pre-determined discrimination threshold 
(e.g. chosen by the user), and $I \in \mathbb{I}$. Assume $I' \in \mathbb{I}$ such that 
there exists a non-empty set $Q \subseteq P_{disc}$ and for all $q \in Q$, $I_q \ne I'_{q}$ 
and for all $p \in P \setminus Q$, $I_p = I'_{p}$. If $\left | f(I) - f(I') \right | > \gamma$, 
then $I$ is called a discriminatory input of the classifier $f$ and is an instance that 
manifests the violation of (individual) fairness in $f$. 
} 
\end{definition}

\theoremstyle{definition}
\begin{definition}{\textbf{(Perturbation)}}
\label{def:perturbation}
{
We define perturbation $g$ as a function 
$g: \mathbb{I} \times \left ( P \setminus P_{disc} \right ) \times \Gamma \rightarrow \mathbb{I}$ 
where $\Gamma = \{-1,+1\}$ captures the set of directions to perturb an input parameter. 
If  $I' = g(I, p, \delta)$ where $I \in \mathbb{I}$, $p \in P \setminus P_{disc}$ and $\delta \in \Gamma$, 
then $I'_p = I_{p} + \delta$ and for all $q \in P \setminus \{p\}$, we have $I'_q = I_{q}$.}
%
\end{definition}
It is worthwhile to mention that the set of directions to perturb an input parameter, i.e. 
$\Gamma$ can easily be extended with more possibilities to perturb. Besides, it can also 
be customized with respect to different input parameters. However, for the sake of brevity, 
we will stick with the simplified version stated in Definition~\autoref{def:perturbation}.

\begin{figure}[t]
\begin{center}
\begin{tabular}{c}
\rotatebox{0}{
\includegraphics[scale = 0.4]{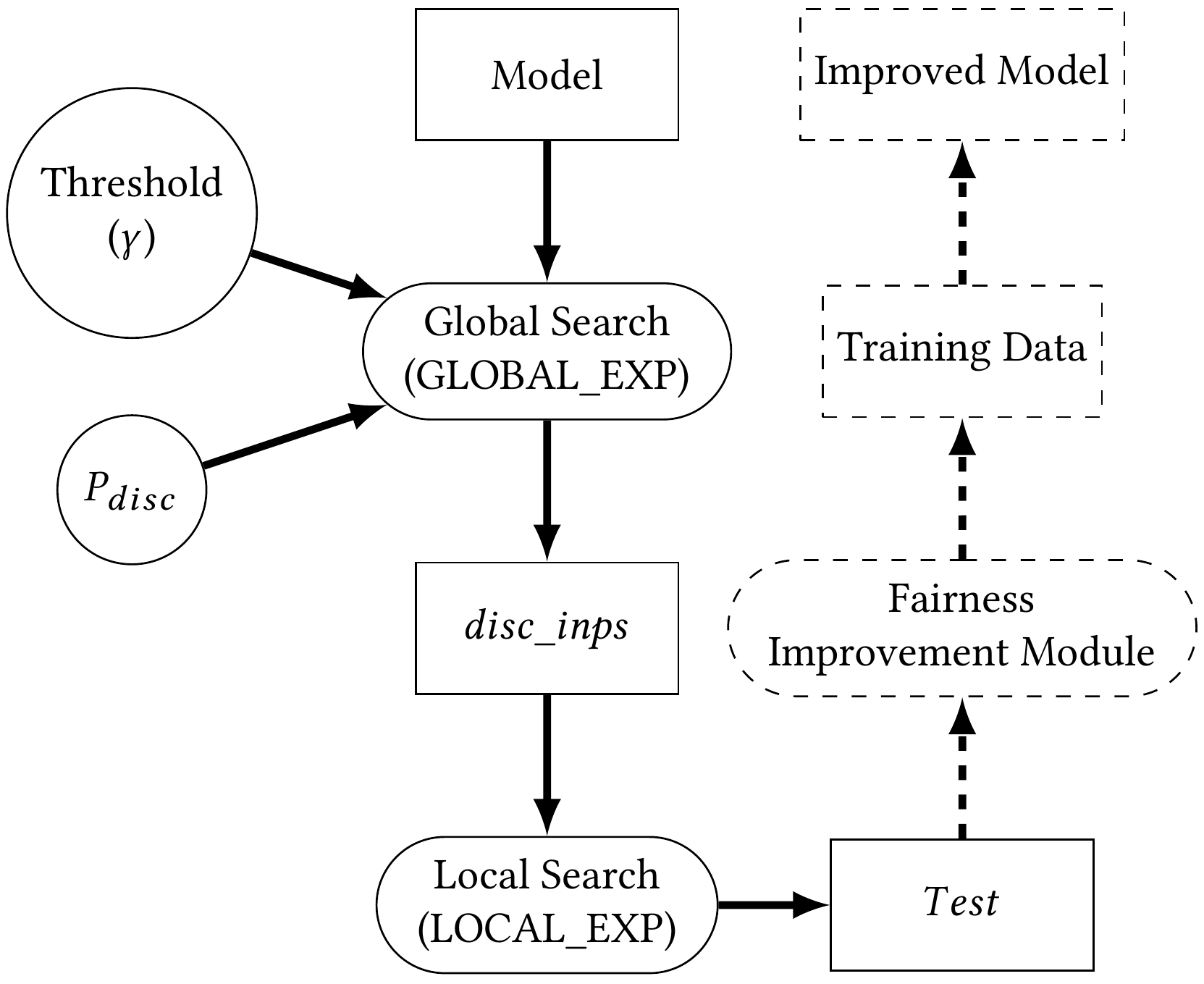}}
\end{tabular}
\end{center}
\vspace*{-0.2in}
\caption{An overview of our \FT approach}
\label{fig:approach}
\vspace*{-0.1in}
\end{figure}

An overview of our overall approach appears in \autoref{fig:approach}. The main contribution 
of this paper is an automated test generator to discover fairness violation. This involves 
two stages: 1) global search ($\mathit{GLOBAL\_EXP}$) and 2) local search 
($\mathit{LOCAL\_EXP}$) over the input domain $\mathbb{I}$. Optionally, the generated test 
inputs can be leveraged to retrain the model under test and improve fairness. 

In the following, we will describe the crucial components of our \FT approach, as shown 
in \autoref{fig:approach}.

\subsection{{Global Search}}
The motivation behind our global search (cf. procedure \textsc{global\_exp} 
in Algorithm~\ref{algo:basic}) is to discover some points in $\mathbb{I}$ that 
can be used to drive our local search algorithm.  
To this end, we first select an input $I$ randomly from the input domain. 
Input $I$, then, is used to generate a set of inputs that cover all possible values 
of sensitive parameters $P_{disc} \subseteq P$. This leads to a set of inputs 
$\mathbb{I}^{(d)}$. We note that the set of sensitive parameters 
(e.g. race, religion, gender) $P_{disc}$ typically has a small size. Therefore, 
despite the exhaustive nature of generating $\mathbb{I}^{(d)}$, this is practically 
feasible. Finally, we discover the discriminatory inputs 
(cf. Definition~\ref{def:discriminatory}) within $\mathbb{I}^{(d)}$ and 
use the resulting discriminatory input set for further exploration during 
our local search over $\mathbb{I}$.

\input{algorithms}

\subsection{{Local Search}}
\label{sec:local-search}
In this test generation phase, we take the inputs generated by our global 
search (i.e. $disc\_inputs$) and then search in the neighbourhood of 
$disc\_inputs$ to discover other inputs with similar characteristics 
(cf. procedure \textsc{local\_exp} in Algorithm~\ref{algo:local}). 
Our search strategy is motivated from the robustness property inherent in common 
machine-learning models. According to the notion of robustness, 
the neighbourhood of an input should produce similar output. 
Therefore, it becomes logical to search the neighbourhood of $disc\_inputs$, 
as these are the discriminatory inputs and their neighbourhood are likely 
to be discriminatory for robust models. 


To search the neighbourhood of $disc\_inputs$, \FT perturbs an input 
$I \in disc\_inputs$ by changing the value of some parameter 
$p \in P \setminus P_{disc}$ (i.e. $I_p$). The value of the parameter $p$ is 
perturbed by $\delta \in \{-1, +1\}$. 
We note that as a side-effect of changing $I_p$, input $I$ is automatically 
modified. This modified version of $I$ is further perturbed in subsequent 
iterations of the inner loop in Algorithm~\ref{algo:local}.
Our \FT approach chooses a parameter $p \in P \setminus P_{disc}$ with probability 
$\sigma_{pr}[p]$ (cf. Algorithm~\ref{algo:local}). For all 
$p \in P \setminus P_{disc}$, initially $\sigma_{pr}[p]$ was assigned 
to $\frac{1}{\left | P \setminus P_{disc} \right |}$. Once $p$ is 
chosen its value is perturbed by $\delta = -1$ with probability 
$\sigma_{v}[p]$ and by $\delta = +1$ with probability $1-\sigma_{v}[p]$. 
$\sigma_{v}[p]$ is initialized to 
$0.5$ 
for all parameters in $p \in P \setminus P_{disc}$.


\FT employs three different strategies, namely \FT random, \FT semi-directed 
and \FT fully-directed, to update the probabilities in $\sigma_{pr}$ and 
$\sigma_{v}$. This is to direct the test generation process with a focus on 
discovering discriminatory inputs. In the following, we will outline the 
different strategies implemented within \FT.  


\subsubsection*{\textbf{\FT random}}
\FT random does not update the initial probabilities assigned to $\sigma_{pr}$ and 
$\sigma_{v}$. This results in $\delta$ (i.e. perturbation value) and $p$ (i.e. the 
parameter to perturb) both being chosen randomly. Intuitively, \FT random explores 
inputs around the neighbourhood of $disc\_inputs$ (i.e. set of discriminatory inputs 
discovered via global search) uniformly at random. Nevertheless, \FT random empirically  
outperforms a purely random search over the input space. This is because it still 
performs a random search in a constrained input region -- specifically, the input 
region that already contains discriminatory inputs. 


\begin{algorithm}[t]
	\caption{\FT semi-directed update probability}
	\label{algo:semidir}
{\small
	\begin{algorithmic}[1]
	 \Procedure{update\_prob}{$I$, $p$, $\mathit{Test}$, $\delta$, $\Delta_v$, $\Delta_{pr}$}
	 	\If {($I \in \mathit{Test} \wedge \delta=-1$) $\vee$ ($I \notin \mathit{Test} \wedge \delta=+1$)}
	 		\State $\sigma_{v}[p] \gets$  $\min$($\sigma_{v}[p] + \Delta_v, 1$)
	 	\EndIf	
	 	\If {($I \notin \mathit{Test} \wedge \delta=-1$) $\vee$ ($I \in \mathit{Test} \wedge \delta=+1$)}
	 		\State $\sigma_{v}[p] \gets$  $\max$($\sigma_{v}[p] - \Delta_v, 0$)
	 	\EndIf
	 \EndProcedure
	\end{algorithmic}}
\end{algorithm}

\begin{algorithm}[t]
	\caption{\FT fully-directed update probability}
	\label{algo:fulldir}
{\small
	\begin{algorithmic}[1]
	\Procedure{update\_prob}{$I$, $p$, $\mathit{Test}$, $\delta$, $\Delta_v$, $\Delta_{pr}$}
	 	\If {($I \in \mathit{Test} \wedge \delta=-1$) $\vee$ ($I \notin \mathit{Test} \wedge \delta=+1$)}
	 		\State $\sigma_{v}[p] \gets$  $\min$($\sigma_{v}[p] + \Delta_v, 1$)
	 	\EndIf	
	 	\If {($I \notin \mathit{Test} \wedge \delta=-1$) $\vee$ ($I \in \mathit{Test} \wedge \delta=+1$)}
	 		\State $\sigma_{v}[p] \gets$  $\max$($\sigma_{v}[p] - \Delta_v, 0$)
	 	\EndIf
	 	\If {$I \in \mathit{Test}$}
	 		\State $\sigma_{pr}[p] \gets$ $\sigma_{pr}[p] + \Delta_{pr}$
	 		\State $\sigma_{pr}[p] \gets$ $\frac{\sigma_{pr}[p]}{\sum_{x \in P \setminus P_{disc}} \sigma_{pr}[x]}$ 
	 		for all $p \in P \setminus P_{disc}$
	 	\EndIf
	 \EndProcedure

	\end{algorithmic}}	
\end{algorithm}

\subsubsection*{\textbf{\FT semi-directed}}
\FT semi-directed drives the test generation by systematically updating 
$\sigma_{v}$, i.e., the probabilities to perturb the value of an input 
parameter by $\delta = -1$ (cf. Algorithm~\ref{algo:semidir}). 
The parameter $p$, to perturb, is still chosen randomly. 
Initially, we choose $\delta \in \{-1, +1\}$  
where the probability that $\delta = -1$ is $\sigma_v[p]$ and the probability 
that $\delta = +1$ is $1 - \sigma_v[p]$. 
If the perturbed input is discriminatory (cf. Definition~\ref{def:discriminatory}), 
then we increase the probability associated with $\sigma_v[p]$ by a pre-determined 
offset $\Delta_v$. Otherwise, $\sigma_v[p]$ is reduced by the same offset $\Delta_v$. 
Intuitively, the updates to probabilities in $\sigma_v$ prioritise a direction 
$\delta \in \{-1,+1\}$ when the respective direction results in discriminatory 
inputs.  


\subsubsection*{\textbf{\FT fully-directed}}
%
\FT fully-directed extends \FT semi-directed by systematically updating the probabilities 
to choose a parameter for perturbation. To this end, we update probabilities in 
$\sigma_{pr}$ during the test generation process (cf. Algorithm~\ref{algo:fulldir}). 
Assume we pick a parameter $p \in P \setminus P_{disc}$ to 
perturb. Initially, we have 
$\sigma_{pr}[p] = \frac{1}{\left | P \setminus P_{disc} \right |}$. If the perturbation 
of the given parameter $p$ by $\delta$ results in a discriminatory input, then we add 
a pre-determined offset $\Delta_{pr}$ to $\sigma_{pr}[p]$. To reflect this change in probability, 
we normalize $\sigma_{pr}[p']$ to 
$\frac{\sigma_{pr}[p']}{\sum_{x \in P \setminus P_{disc}} \sigma_{pr}[x]}$ for 
every $p' \in P \setminus P_{disc}$. Intuitively, the updates to probabilities in 
$\sigma_{pr}$ prioritize a parameter when perturbing the respective parameter 
results in discriminatory inputs.


%

\subsection{{Estimation using LLN}}
\label{sec:estimation-lln}
An attractive feature of \FT is that we can estimate the percentage of discriminatory 
inputs in $\mathbb{I}$ for any given model. We leverage the Law of Large Numbers (LLN) 
from probability theory to accomplish this. Let $\Lambda$ be an experiment. In this 
experiment, we generate  $m$ inputs uniformly at random. These are independent and 
identically distributed (IID) samples $I_1, I_2 \ldots I_m$. We execute these inputs 
and count the number of inputs that are discriminatory in nature. Let $m^{'}$ be the 
number of inputs that are discriminatory. $\Lambda$ then outputs the percentage 
$\overline{m} = \frac{m^{'} \times 100}{m}$.

$\Lambda$ is conducted $K$ times. In each instance of the experiment, we collect the 
outcome $\overline{m}_1, \overline{m}_2 \ldots \overline{m}_{K}$. 
Let $\overline{M} = K^{-1} \sum_{i = 1}^{K} \overline{m}_i$. According to LLN,  the 
average of the results, i.e. $\overline{M}$, obtained from a large number of trials, 
should be close to the expected value, and it will tend to become closer as more 
trials are performed. This implies as, 
\begin{gather*}
K \to \infty  \\
\overline{M} \to M^{*}
\end{gather*}
where $ M^{*}$ is the true percentage of the discriminatory inputs present in $\mathbb{I}$ 
for the machine-learning model under test. This phenomenon was observed in our experiments. 
\autoref{fig:example}(b) shows that the $\overline{M}$ converges only after 400 trials 
(i.e. $K=400$).

\subsection{{Improving Model Fairness}}
\label{sec:retrain}

%
It has been observed that generated test inputs showing the violation 
of desired-properties in machine-learning models can be leveraged for 
improving the respective properties. 
This was accomplished via augmenting 
the training dataset with the generated test inputs and retraining the 
model~\cite{deeptest}. 

Hence, we intend to evaluate the usefulness of our generated test inputs 
to improve the model fairness via retraining. 
To this end, \FT has a completely automated module that guarantees reduction 
of the percentage of discriminatory inputs in $\mathbb{I}$. We achieve 
this by systematically adding portions of generated discriminatory 
inputs to the training dataset. 



\begin{algorithm}[t]
	\caption{Retraining}
    \label{algo:retraining}
    {\small
    \begin{algorithmic}[1] 
        \Procedure{Retraining}{$f$, $\mathit{Test}$, $\mathit{training\_data}$} 
            \State $N \gets \infty$ 
            \State $f_{cur} \gets f$ 
            \For{$i $ in (2, N)}
            	\State $p_i \gets $ a random real number between $(2^{i - 2},  2^{i-1})$
            	\If {$p_i > 100$}
            		\State Exit the loop
            	\EndIf
            	\State $k \gets$ len($\mathit{training\_data}$)
            	\State $n_{addn} \gets \frac{p_i \cdot k}{100}$ 
            	\State $\mathit{TD_{addn}} \gets$ randomly selected $n_{addn}$ inputs from $Test$ 
            	\State $\mathit{TD_{new}} \gets \mathit{training\_data} \cup \mathit{TD_{addn}}$
            	
            	\State $f_{new} \gets $ model trained using $\mathit{TD_{new}}$
            	\LineComment {Estimate the number of discriminatory inputs (section~\ref{sec:estimation-lln})}
            	\State $\mathit{{fair}_{cur}} \gets $ LLN\_Fairness\_Estimation ($f_{cur}$)
            	\State $\mathit{{fair}_{new}} \gets $ LLN\_Fairness\_Estimation ($f_{new}$)
            	\If {($\mathit{{fair}_{cur}} > \mathit{{fair}_{new}}$)}
         			\State $f_{cur} \gets f_{new}$ 
          	 	\Else 
          	 		\State Exit the loop
          	 	\EndIf
            \EndFor
           \Return $f_{cur}$
            
        \EndProcedure
    \end{algorithmic}}
 \end{algorithm}


Assume $\mathit{Test}$ be the set of discriminatory inputs generated by 
\FT. \FT is effective in generating discriminatory inputs and the size 
of the set $\mathit{Test}$ is usually large. A naive approach to retrain 
the model will be to add all generated discriminatory inputs to the 
training dataset. Such an approach is likely to fail to improve the 
fairness of the model. 
This is because the generated test inputs are targeted towards finding 
discrimination and are unlikely to follow the true distribution of the 
training data. Therefore, blindly adding all the test inputs to the 
training set will bias its distribution towards the distribution of our 
generated test inputs. 
%
To solve this challenge, it is important that only 
portions of discriminatory inputs from $\mathit{Test}$ are added to 
the training dataset. 

Let $p_i$ be the percentage, with respect to the size of the training 
data, that we choose at any given iteration $i$. If size of training 
data is $M$, then we select $\frac{p_i \cdot M}{100}$ discriminatory 
inputs from $\mathit{Test}$ at random and add these discriminatory inputs 
to the training dataset. 
For $i \in [2,N]$, we set $p_i$ randomly in a range between $\mathopen[ 2^{i-2}, 2^{i-1} \mathclose]$. 
The intuition behind this is to find an efficient mechanism to systematically 
add inputs from $\mathit{Test}$ to the training dataset and to approximate 
the optimal reduction in discriminatory inputs. We terminate the 
process when adding inputs from $\mathit{Test}$ to the training dataset does 
not decrease the estimated fraction of discriminatory inputs in $\mathbb{I}$.  
The currently trained model (i.e. $f_{cur}$ in Algorithm~\ref{algo:retraining}) 
is then taken as the improved model with better (individual) fairness score.
In this way, we can guarantee that our retraining process always terminates  
with a reduction in discriminatory inputs.

Our retraining strategy is designed to be fast without sacrificing the fairness 
significantly. Our main objective is to demonstrate that \FT generated test 
inputs can indeed be used by the developers to improve the individual fairness of  
their models. 
The amount of added test inputs (generated by \FT) is chosen from exponentially 
increasing intervals (i.e. the interval $[2^{i-2}, 2^{i-1}]$ in 
Algorithm~\ref{algo:retraining}). Such a strategy is taken to quickly scope the 
sensitivity of the model with respect to the generated test data. 
%
%
Moreover, by choosing a random number $p_i$ in the interval, we try not to 
overshoot the value of $p_i$ by a large margin that causes the optimal 
reduction of discriminatory inputs in $\mathbb{I}$. As a result, our proposed 
retraining strategy maintains a balance between improving model fairness and 
the efficiency of retraining.

It is well known that adding more data to a machine-learning algorithm is likely 
to lead to increased accuracy~\cite{data_effectiveness}. A relevant challenge here 
is attributed to the labeling of the generated test data. 
There exists a number of effective strategies to tackle this problem. One such strategy 
is finding the label via a simple majority of a number of classifiers~\cite{majority_voting}. 
Majority voting has been shown to be very effective for a wide range of problems~\cite{deepxplore} 
and we believe it should be readily applicable in our context of improving fairness 
as well. Nevertheless, test data labeling is an orthogonal problem in the domain of 
machine learning and we consider it to be beyond the scope of the problem 
targeted by \FT.


\subsection{{Termination}}
\FT can be configured to have various termination conditions depending on the particular 
use case of the developer. In particular, \FT can be terminated with the following 
possible conditions: 
\begin{enumerate}
\item \FT can terminate after it has generated a user specified number of discriminatory 
inputs from $\mathbb{I}$. This feature can be used when a certain number of discriminatory 
inputs need to be generated for testing, evaluation or retraining of the model.

\item \FT can also terminate within a given time bound. This is useful to quickly check if 
the model exhibits discrimination for a particular set of sensitive parameters. 

\end{enumerate}
In our evaluation, we used both the termination criteria to evaluate the effectiveness 
and efficiency of \FT.

%% file: algorithms.tex
\algdef{SE}[VARIABLES]{Variables}{EndVariables}
   {\algorithmicvariables}
   {\algorithmicend\ \algorithmicvariables}
\algnewcommand{\algorithmicvariables}{\textbf{global variables}}

\algnewcommand{\LineComment}[1]{\State\(\triangleright\) #1}
\let\oldReturn\Return
\renewcommand{\Return}{\State\oldReturn}

\begin{algorithm}[t]
	
    \caption{Global Search}
    \label{algo:basic}
    {\small
    \begin{algorithmic}[1] 
        \Procedure{global\_exp}{$P$, $P_{disc}$} 
            \State $disc\_inps \gets$  $\phi$
            \LineComment {\textsf{$N$ is the number of trials in global search}}
            \For{$i $ in (0, N)}
            	\State Select an input $I \in \mathbb{I}$ at random
            	\LineComment {\textsf{$\mathbb{I}^{(d)}$ extends $I$ with all possible values of $P_{disc}$}}
            	\State $\mathbb{I}^{(d)} \gets$ $\{I'\ |\ \forall p \in P \setminus P_{disc}.\ I_p = I'_{p}\}$
            	 \If {($\exists I, I' \in \mathbb{I}^{(d)}$, $\left | f(I) - f(I') \right | > \gamma$)}
            		\State $disc\_inps \gets$  $disc\_inps \cup \{I\}$
            	\EndIf
            \EndFor
           \Return $disc\_inps$
            
        \EndProcedure
    \end{algorithmic}}
 \end{algorithm}   
    
 \begin{algorithm}[t]
    \caption{Local Search}
    \label{algo:local}
    {\small
    \begin{algorithmic}[1]
    \Procedure{local\_exp}{$disc\_inps$, $P$, $P_{disc}$, $\Delta_v$, $\Delta_{pr}$}
    \State $\mathit{Test} \gets$  $\phi$
    \State Let $P'= P \setminus P_{disc}$
    \State Let $\sigma_{pr}[p] = \frac{1}{\left | P' \right |}$ for all $p \in P'$
    \State Let $\sigma_{v}[p] = 0.5$ for all $p \in P'$ 
    \For{$I \in disc\_inps$}
    	\LineComment {\textsf{$N$ is the number of trials in local search}}
    	\For{$i $ in (0, N)}
    		\State Select $p \in P'$ with probability $\sigma_{pr}[p]$
    		\State Select $\delta = -1$ with probability $\sigma_{v}[p]$
    		\LineComment {\textsf{Note that $I$ is modified as a side-effect of modifying $I_p$}}
    		\State $I_p \gets I_p + \delta$
    		\LineComment {\textsf{$\mathbb{I}^{(d)}$ extends $I$ with all values of $P_{disc}$}}
        \State $\mathbb{I}^{(d)} \gets$ $\{I'\ |\ \forall p \in P \setminus P_{disc}.\ I_p = I'_{p}\}$
        \If {($\exists I, I' \in \mathbb{I}^{(d)}$, $\left | f(I) - f(I') \right | > \gamma$)}
        		\LineComment {\textsf{Add the perturbed input $I$}}
            \State $\mathit{Test} \gets$ $\mathit{Test} \cup \{I\}$ 
        \EndIf
        \State update\_prob($I$, $p$, $\mathit{Test}$, $\delta$, $\Delta_v$, $\Delta_{pr}$)
    	\EndFor
    \EndFor
    \Return $\mathit{Test}$
    \EndProcedure
    \end{algorithmic}}
\end{algorithm}

%% file: results.tex
\section{Results}
\label{sec:results}

\subsubsection*{\textbf{Experimental setup}}

We evaluate \FT across a wide variety of classifiers, including a classifier 
which was designed to be fair. Some salient features of these classifiers 
are outlined in \autoref{classifiers}. In particular, Fair SVM (cf. \autoref{classifiers}) 
was specifically designed with fairness in mind~\cite{zafar_fair_classifier}. 
The rest of the classifiers under test are the standard implementations found 
in Python's Scikit-learn machine learning library. These classifiers are 
used in a wide variety of applications by machine-learning engineers across 
the world.

Other than Fair SVM ~\cite{zafar_fair_classifier}, we have used Scikit-learn's 
Support Vector Machines (SVM), Multi Layer Perceptron (MLPC), Random Forest 
and Decision Tree implementations for our experiments. We also evaluate an 
Ensemble Voting Classifier (Ensemble), in which we take the combination of 
two classifier predictions. The classifiers we use are Random Forest and 
Decision Tree estimators (cf. \autoref{classifiers}).

\vspace*{-0.1in}
\begin{table}[h]
\caption{Subject classifiers used to evaluate \FT}
\vspace*{-0.2in}
\label{classifiers}
\begin{center}
\begin{tabular}{| c | c | c |}
\cline{1-3}
Classifier name & Lines of python code & Input domain \\ \cline{1-3}
       Fair SVM~\cite{zafar_fair_classifier}   &     913          &          \multirow{6}{*}{$10^6$}    \\
       SVM  &         1123      &              \\
       MLPC  &         1308      &              \\
       Random Forest   &   1951           &              \\
       Decision Tree    &    1465           &              \\ 
       Ensemble    &      3466         &              \\ \hline
\end{tabular}
\end{center}
\vspace*{-0.1in}
\end{table}


All classifiers listed in \autoref{classifiers} are used for predicting the 
income. 
These classifiers are trained with the data obtained from the US census~\cite{us_census}. 
The size of this training data set is around 32,000. We train all the six 
classifiers on this training data. The objective is to classify whether 
the income of an individual is above \$50,000 (captured via classifier 
output ``+1") or below (captured via classifier output ``-1"). For all 
the classifiers, set of discriminatory parameters, i.e. $P_{disc}$ is the 
{\em gender} of an individual. The threshold value for identifying 
a discriminatory input is set to zero. 
This means, if $I$ differs from $I'$ only in being $\mathit{Gender}_A$ or 
$\mathit{Gender}_B$, then $I$ or $I'$ are discriminatory inputs of a 
classifier $f$ when $\left | f(I) - f(I') \right | \geq 0$.
In our experiments we set the perturbation $\delta \in \{-1, + 1 \}$ 
and both $\Delta_{v}$ and $\Delta_{pr}$ as $0.001$. These are user 
defined variables that guide our \FT approach. In particular, these 
variables are used to systematically refine the probabilities to choose 
an input parameter to perturb and to choose a perturbation value 
$\delta$ (cf. \autoref{sec:approach}). 


We implement \FT in Python, as it is a popular choice of language for 
the development of machine-learning models and related applications. 
The implementation is around 600 lines of python code. 
All our experiments were performed on an Intel \texttt{i7} processor 
having 64GB of RAM and running Ubuntu 16.04.


\subsubsection*{\textbf{Key results}}
\begin{figure}[t]
\begin{center}
\rotatebox{0}{
\includegraphics[scale = 0.31]{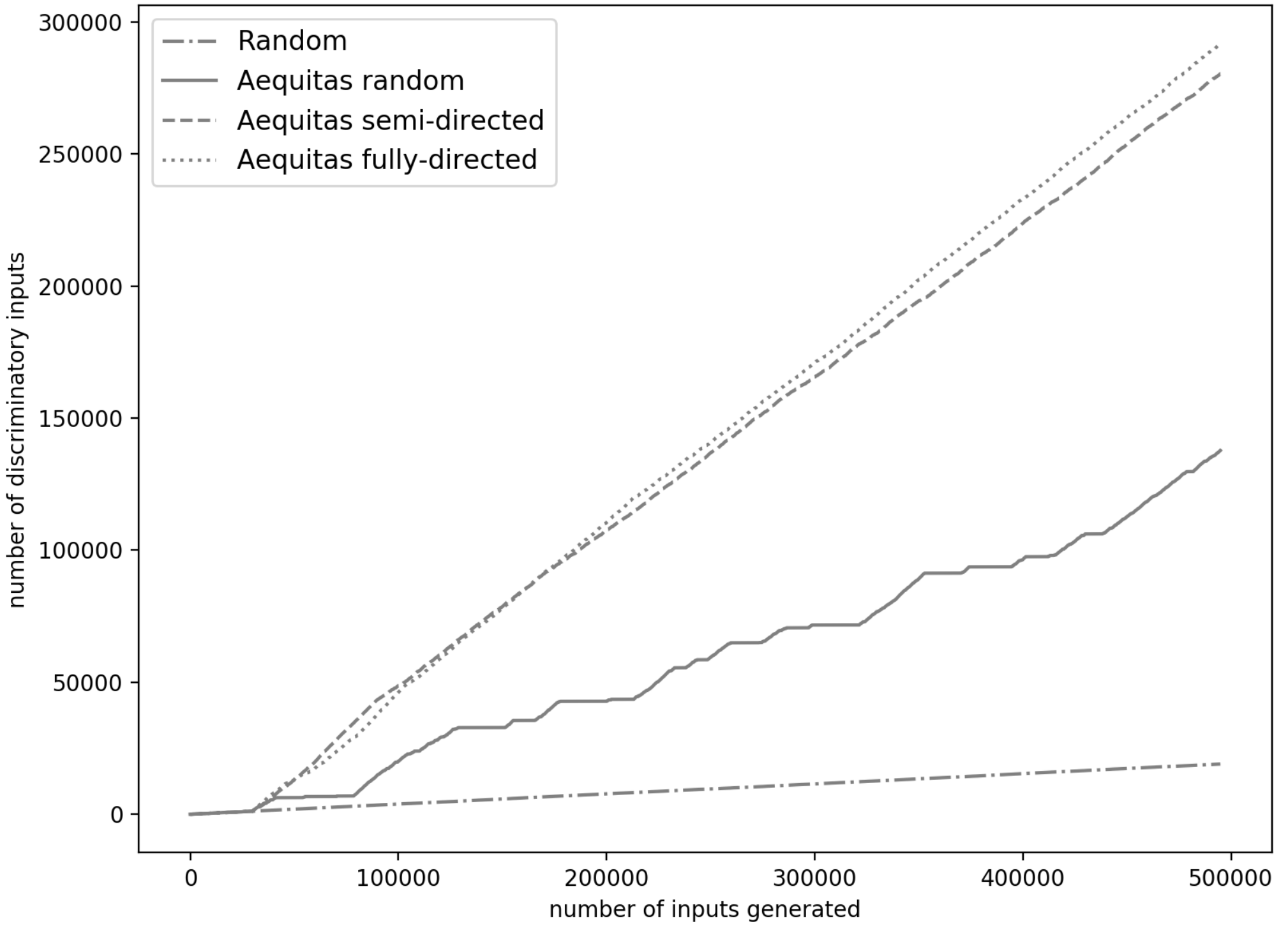}}
\end{center}
\vspace*{-0.1in}
\caption{The effectiveness of \FT}
\label{fig:effectivenessGraph}
\vspace*{-0.1in}
\end{figure}

We use three different test generation methodologies, namely 
\FT random, \FT semi-directed and \FT fully-directed. These 
methodologies differ with respect to the increasing levels 
of sophistication in systematically searching the input space 
(cf. \autoref{sec:local-search}). 
In particular, \FT fully-directed involves the highest level 
of sophistication in searching the input space. 
As expected, \FT fully-directed consistently outperforms the 
\FT random and \FT semi-directed, as observed from 
\autoref{fig:effectivenessGraph}. However, \FT fully-directed  
and \FT semi-directed demand more computational resources 
per unit time than \FT random. As a result, \FT random is more 
appropriate to use, as compared to the rest of our approaches, 
for testing with limited computational resources per unit time.
The test subject used in \autoref{fig:effectivenessGraph} 
was the Fair SVM (cf. \autoref{classifiers}).


To illustrate the power of our \FT approach over the 
state-of-the-art fairness testing~\cite{fairness_fse_2017}, 
we also compare our approaches with the state-of-the-art, 
which, in turn is captured via ``Random" in 
\autoref{fig:effectivenessGraph}. It is evident that even 
the least powerful technique implemented within our \FT approach 
(i.e. \FT random) significantly outperforms the 
state-of-the-art. 
In our evaluation, we discovered that \FT is more effective than the 
state-of-the-art random testing by a factor of 9.6 on average and up to a factor 
of 20.4. We measured the effectiveness via the number of discriminatory 
inputs generated by a test generation technique. \FT also provides 
capabilities to automatically retrain a machine-learning model with the 
objective to reduce the number of discriminatory inputs. To this end, 
\FT reduced the number of discriminatory inputs by 43.2\% on average 
with a maximum reduction of 94.36\%.

\begin{center}
\begin{tcolorbox}[width=\columnwidth, colback=gray!25,arc=0pt,auto outer arc]
\textbf{RQ1: How effective is \FT in finding discriminatory inputs?}
\end{tcolorbox}
\end{center}

We evaluate the capability of \FT in effectively generating discriminatory inputs. 
For all the subject classifiers, we measure the effectiveness of our test 
algorithms via the number of discriminatory inputs generated with respect to the 
number of total inputs generated. 

A purely random approach is not effective in generating discriminatory inputs. 
As observed from \autoref{fig:effectivenessGraph}, the number of discriminatory 
inputs generated by such an approach does not increase rapidly over the number 
of inputs generated. This is expected, as a purely random approach does not 
incorporate any systematic strategy to discover inputs violating fairness. The 
ineffectiveness of random testing persists across all the subject classifiers, 
as observed in \autoref{effectiveness}. 


As observed from \autoref{effectiveness}, all test generation approaches 
implemented within \FT outperform a purely random approach. In particular, 
the rate at which our \FT approach generates discriminatory inputs is 
significantly higher than a purely random approach. As a result, \FT 
provides scalable and effective technique for machine learning engineers 
who aim to rapidly discover fairness issues in their models. \FT random, 
\FT semi-directed and \FT fully-directed involve increasing level of 
sophistication in directing the test input generation. 
As a result, \FT fully-directed approach performs the best among all our 
test generators. In particular, \FT semi-directed is on an average 46.7\% 
and up to 64.9\% better than \FT random. Finally, \FT full-directed is 
on an average 29.5\% and up to 56.56\% better than \FT semi-directed. 

By design, \FT does not generate any false positives. This means that 
any discriminatory input generated by \FT are indeed discriminatory to 
the model under test, subject to the chosen threshold of discrimination. 

\begin{table*}[h]
	\centering
	\caption{Effectiveness of \FT approach}
	\vspace*{-0.15in}
	\label{effectiveness}
	\begin{tabular}{| c | c | c | c | c | c | c | c |}
	\cline{1-8}
	Classifier    &  Random~\cite{fairness_fse_2017} & \multicolumn {2} { c |} {\makecell{\FT \\ random} }& 
	\multicolumn {2} {c |} {\makecell{\FT \\ semi-directed} }& \multicolumn {2} {c |} {\makecell{\FT \\ fully-directed} }\\ \cline{1-8}
	{} 				& \makecell {\% discriminatory \\ input} & \makecell {\% discriminatory \\ input} & \makecell{\# inputs\\ generated}   & \makecell {\% discriminatory \\ input} & \makecell{\# inputs \\generated}   & \makecell {\% discriminatory \\ input} & \makecell{\# inputs \\generated}  \\ \cline{1-8}
	Fair SVM      & 3.45           & 39.4  & 315640   & 65.2 & 322725     & 70.32 & 357375    \\
    SVM           & 0.18           & 0.53  &   54683   & 0.574 &  88095   & 1.22  & 100101    \\
	MLPC          & 0.3466         & 2.15  & 218727    & 2.39 &  129556  & 2.896 &  141666  \\ 			Random Forest & 8.34           & 18.312 & 218727  & 21.722 & 264523  & 34.98 & 282973   \\
	Decision Tree & 0.485          & 2.33   & 153166  & 2.89  & 179364    & 6.653 &  248229  \\ 
	Ensemble      & 8.23           & 22.34 & 187980   & 36.08  & 458910  & 37.9 & 545375 \\ \hline
\end{tabular}
\vspace*{-0.15in}
\end{table*}

\begin{center}
\begin{tcolorbox}[width=\columnwidth, standard jigsaw, opacityback=0, arc=0pt, auto outer arc]
\textbf{Finding:} \FT fully-directed approach outperform a purely random approach up to a factor of 
20.4 in terms of the number of discriminatory inputs generated. It also performs up to 56.7\% 
better than \FT semi-directed, which, in turn performs up to 64.9\% better than \FT random, 
our least sophisticated approach. 
\end{tcolorbox}
\end{center}

\begin{center}
\begin{tcolorbox}[width=\columnwidth, colback=gray!25,arc=0pt,auto outer arc]
\textbf{RQ2: How efficient is \FT in finding discriminatory inputs?}
\end{tcolorbox}
\end{center}

\begin{table}[h]
	\centering
	\caption{Test generation efficiency}
	\vspace*{-0.15in}
	\label{efficiency}
	{\scriptsize
	\begin{tabular}{| c | c | c | c | c |}
	\cline{1-5}
	Classifier    &  Random & \makecell{\FT \\ random} & 
	\makecell{\FT \\ semi-directed}& \makecell{\FT \\ fully-directed} \\ 
	\cline{1-5}
	Fair SVM   & 1589.87s  &      534.47s    &  345.65s    &   228.14s       \\
	SVM        &   7159.54s &       3589.9s    &  2673.8s     &   2190.21s     \\
	MLPC       &   6157.23s &        759.63s  &  431.76s   &   207.87s     \\
	Random Forest & 9563.12s &       2692.98s   &  1334.67s  & 1145.34s      \\
	Decision Tree & 1035.32s &      569.13s    &   371.89s   &   254.25s   \\ 
	Ensemble    &  6368.79s &       2178.45s    &  1067.75s  &   989.43s   \\ \hline
\end{tabular}}
\vspace*{-0.15in}
\end{table}

\autoref{efficiency} summarizes how much time each of the methods takes to generate 
10,000 discriminatory inputs. On an average \FT random performs 64.42\% faster than 
the state of the art. The improvement in \FT fully-directed is even more profound. 
On an average, \FT fully-directed is 83.27\% faster than the state of the art, with 
a maximum improvement of 96.62\% in the case of Multi Layer Perceptron. 

It is important to note that the reported time in \autoref{efficiency} includes both 
the time needed for test generation and for test execution. Hence, the reported time 
is highly dependent on the execution time of the model under test.

\begin{center}
\begin{tcolorbox}[width=\columnwidth, standard jigsaw, opacityback=0, arc=0pt, auto outer arc]
\textbf{Finding:}  \FT fully-directed is 83.27\% faster than the state of the art, with a maximum improvement of 96.62\% in the case of Multi Layer Perceptron. 
\end{tcolorbox}
\end{center}

\begin{center}
\begin{tcolorbox}[width=\columnwidth, colback=gray!25,arc=0pt,auto outer arc]
\textbf{RQ3: How useful are the generated test inputs to improve the fairness of the model? }
\end{tcolorbox}
\end{center}

\vspace*{-0.15in}
\begin{table}[h]
\centering
\caption{Retraining Effectiveness}
\vspace*{-0.15in}
\label{retrain-table}
{\scriptsize
\begin{tabular}{| c | c | c | c | c |}
\cline{1-5}
Classifier    &  \multicolumn {2} { c |} {\makecell{estimated \% of disc input\\ in $\mathbb{I}$  (95\% confidence interval)} } & \makecell{ \%Impr} & \makecell{\%Inps \\ added }\\ \cline{2-3}
& \makecell{before \\ retraining} & \makecell{after \\ retraining} & & \\\cline{1-5}
Fair SVM      & 3.86 (3.76, 3.95)        & 2.89 (2.64, 3.14)  &    25.15     & 15.6                                                \\
SVM           & 0.33 (0.14, 0.51)          & 0.12 (0.09, 0.14)  &       63.54   & 26.9                                                  \\
MLPC          & 0.39 (0.36, 0.42)         & 0.28 (0.27, 0.29)  &      30.12    & 23.7                                                  \\
Random Forest & 8.84 (8.78, 8.91)           & 6.68 (6.35, 7.01)  &     24.48   & 32.4                                                  \\
Decision Tree & 0.48 (0.45, 0.51)        & 0.027 (0.026, 0.028)   &      94.36   & 10.6                                                  \\
Ensemble      & 7.73 (7.14, 8.32)         & 6.06 (5.64, 6.48)      &    21.58    &28.3                                                  \\ \hline
\end{tabular}}
\vspace*{-0.15in}
\end{table}

\FT has a completely automated module which guarantees a decrease in the percentage 
of discriminatory inputs in $\mathbb{I}$. The discriminatory inputs, as discovered 
by \FT, were systematically added to the training dataset (cf. \autoref{sec:retrain}). 
The results of retraining the classifiers appear in \autoref{retrain-table}. 
In general, retraining the classifiers is not significantly time consuming. 
In particular, each classifier was retrained within an hour. For some classifiers, 
such as the SVM, our retraining scheme only took a few minutes. 

We leverage the law of large numbers (LLN) from statistical theory to estimate 
the percentage of discriminatory inputs in $\mathbb{I}$ (cf. \autoref{sec:estimation-lln}). 
In particular, we randomly sample a large number of inputs from $\mathbb{I}$ and compute 
the ratio of discriminatory inputs to the total inputs sampled. This experiment is repeated 
a large number of times and the average of the computed ratio is used as the estimate for 
the percentage of discriminatory inputs in $\mathbb{I}$. We note from statistical 
theory that as the number of experiment is repeated a large number of times, the average 
of the computed ratio should be close to the expected fraction of discriminatory inputs 
in $\mathbb{I}$. We also compute the 95\% confidence interval estimate for the percentage 
of discriminatory inputs in $\mathbb{I}$. 
It is useful to note that these intervals are fairly tight and that adds to the confidence
we have in our point estimates as well.


As observed from \autoref{retrain-table}, \FT is effective in reducing the percentage 
of discriminatory inputs in $\mathbb{I}$ for all the classifiers under test. 
Specifically, we observe an average improvement of 43.2\%, in terms of reducing the 
discriminatory inputs. Using our retraining module, we added an average of only 
7463 datapoints (22.92\% of the original training data) to achieve the 
result obtained in \autoref{retrain-table}.


\begin{center}
\begin{tcolorbox}[width=\columnwidth, standard jigsaw, opacityback=0, arc=0pt, auto outer arc]
\textbf{Finding:} Retraining using \FT lowers the discrimination percentage in $\mathbb{I}$ by an average of 43.2\% and up to 94.36\%.
\end{tcolorbox}
\end{center}


%% file: relatedWork.tex
 \section{Related Work}
 \label{sec:relatedWork}
 
In this section, we review the related literature and position our 
work on fairness testing.  
 
\smallskip\noindent
\textbf{Fair Machine Learning Models}
The machine learning research community have turned their attention 
on designing classifiers that avoid discrimination~\cite{zafar_fair_classifier,fairness_aware_kamishima,Disparate_Impact_Feldman,real_world_goals_goh,fairness_awareness_dwork}.
These works primarily focus on the theoretical aspects of classifier 
models to achieve fairness in the classification process. Such a goal 
is either achieved by pre-processing training data or by modifying 
existing classifiers to limit discrimination. 
Our work is complementary to the approaches that aim to design fair 
machine-learning models. We introduce an efficient way to search the 
input domain of classifiers whose goal is to achieve fairness in decision 
making. We wish to provide a mechanism for these classifiers to quickly 
evaluate their fairness properties and help improve their fairness in 
decision making via retraining, if necessary.

\smallskip\noindent
\textbf{Fairness Testing}
From the software engineering point of view, the research on validating 
the fairness of machine-learning models is still at its infancy. 
A recent work~\cite{fairness_fse_2017} along this line of research defines 
software fairness and discrimination, including a causality-based 
approach to algorithmic fairness. However, in contrast to our \FT approach, 
the focus of this work is more on defining fairness and tests were generated 
in random~\cite{fairness_fse_2017}. 
In particular, \FT can be used as a directed test generation module to 
uncover discriminatory inputs and discovery of these inputs is essential 
to understand individual fairness~\cite{fairness_awareness_dwork} of a 
machine-learning model.
In addition to this and unlike existing approach~\cite{fairness_fse_2017}, 
\FT provides a module to automatically retrain the machine-learning models 
and reduce discrimination in the decisions made by these models. 

\smallskip\noindent
\textbf{Testing and Verification of Machine Learning models}
DeepXplore~\cite{deepxplore} is a whitebox differential testing algorithm 
for systematically finding inputs that can trigger inconsistencies between 
multiple deep neural networks (DNNs). The neuron coverage was used as a 
systematic metric for measuring how much of the internal logic of a DNNs 
have been tested. More recently, DeepTest~\cite{deeptest} leverages 
metamorphic relations to identify erroneous behaviors in a DNN. The usage 
of metamorphic relations somewhat solves the limitation of differential 
testing, especially to lift the requirement of having multiple DNNs implementing 
the same functionality. Finally, a feature-guided black-box approach is proposed 
recently to validate the safety of deep neural networks~\cite{marta_testing}. 
This work uses their method to evaluate the robustness of neural networks 
in safety-critical applications such as traffic sign recognition. 

The objective of these works, as explained in the preceding paragraph, 
is largely to evaluate the robustness property of a given machine-learning model. 
In contrast, we are interested in the {\em fairness property}, 
which is fundamentally different from robustness. Therefore, validating fairness 
requires special attention along the line of systematic test generation. 


\smallskip\noindent
\textbf{Search based testing}
Search-based testing has a long and varied history. The most common techniques 
are hill climbing, simulated annealing and genetic algorithms~\cite{search_based_testing_survey}. 
These have been applied extensively to test applications that largely fall in the 
class of deterministic software systems. \FT is the first instance in our knowledge 
that employs a novel search algorithm to test the fairness of machine-learning 
systems. We believe that we can port \FT for the usage in a much wider 
machine-learning context.

%% file: threatsToValidity.tex
\section{Threats to Validity}
\label{sec:threats}

The effectiveness and efficiency of \FT critically depends on the following factors: 


\smallskip\noindent
\textbf{Robustness}:
Our \FT approach is based on the hypothesis that 
the machine-learning models under test exhibit robustness. 
This is a reasonable assumption, as we expect the models under test 
to be deployed in production settings. 
%
%
As evidenced by our evaluation, \FT approach, which 
is based on the aforementioned hypothesis, was effective to localize the 
search in the vicinity of discriminatory input regions for state-of-the-art 
models.

\smallskip\noindent
\textbf{Training data and access to model}: \FT needs access to the 
training data and the training mechanism of the machine-learning model to 
be able to evaluate and retrain the model. 
Without access to the training data, \FT will not 
be able to successfully improve the fairness of the model. This is because 
\FT is used to generate test inputs that violate fairness and augment the 
original training set to improve the model under test. The generated test 
inputs, however, is not sufficient to train a machine-learning model from 
scratch. 


\smallskip\noindent
\textbf{Input Structure}: \FT works on real-valued inputs. \FT, in 
its current form, does not handle image, sound or video inputs. This, 
however, does not diminish the applicability of \FT. Numerous 
real-world applications still use only real-valued data for prediction. 
These include applications in finance, security, social welfare, education, 
healthcare and human resources. 
Examples of applications include income prediction, crime prediction, 
disease prediction, job short-listing and college short-listing, among others. 
For models that take inputs such as images and videos, we need to 
incorporate additional techniques for automatically generating valid input data. 
However, we believe that the core idea behind our \FT approach, namely the global 
and the local search employed over the input space, will still remain valid. 

\smallskip\noindent
\textbf{Probability change parameter}: 
The users of \FT will have to experiment and carefully choose 
$\Delta_v$ and $\Delta_{pr}$ 
values which change the probabilities of choosing $p$ (i.e. the input 
parameter to perturb) and $\delta$ (i.e. the perturbation value). 
If $\Delta_v$ (respectively, $\Delta_{pr}$) is too high, then an 
overshoot might occur and a certain discriminatory input region may 
never be explored. If $\Delta_v$ (respectively, $\Delta_{pr}$) is 
too low, then the effectiveness of \FT semi-directed and \FT fully-directed 
would be very similar to \FT random. In our experiments, we evaluated 
with a few $\Delta_v$ and $\Delta_{pr}$ values before our results stabilized.

\smallskip\noindent
\textbf{Limited discriminatory input features}: We evaluate \FT with discriminatory 
input feature {\em gender}. Hence, we cannot conclude the effectiveness of 
\FT for other potentially discriminatory input features. However, the mechanism 
behind \FT is generic and allows extensive evaluation for other discriminatory 
input features in a future extension of the tool.



%% file: conclusion.tex
\section{Conclusion}
\label{sec:conclusion}

In this paper, we propose \FT -- a fully automated and directed 
test generation strategy to rapidly generate discriminatory 
inputs in machine-learning models. The key insight behind \FT 
is to exploit the robustness property of common machine learning 
models and use it to systematically direct the test generation 
process. \FT provides statistical evidence on the number of 
discriminatory inputs in a model under test. Moreover, \FT 
incorporates strategies to systematically leverage the generated 
test inputs to improve the fairness of the model. We evaluate 
\FT with state-of-the-art classifiers and demonstrate that \FT 
is effective in generating discriminatory test inputs as well 
as improving the fairness of machine-learning models. 
At its current state, however, \FT does not have the 
capability to localize the cause of discrimination in a model. 
Further work is required to isolate the cause of discrimination 
in the model.

\FT provides capabilities to lift the state-of-the-art in 
testing machine-learning models. We envision to extend our 
\FT approach beyond fairness testing and for machine-learning 
models taking complex inputs including images and videos. 
We hope that the central idea behind our \FT approach would 
influence the rigorous software engineering principles and 
help validate machine-learning applications used in sensitive 
domains. For reproducibility and advancing the state of research, 
we have made our tool and all experimental data publicly available: 
\begin{center}
\url{https://github.com/sakshiudeshi/Aequitas}
\end{center}